\documentclass{article} % For LaTeX2e
\usepackage[final]{colm2026_conference}
\usepackage[T1]{fontenc}
\usepackage{microtype}
\usepackage{hyperref}
\usepackage{url}
\usepackage{booktabs}

\usepackage{tcolorbox}
\usepackage{graphicx}
\usepackage{tikz}
\usetikzlibrary{shapes, fit}
\usepackage{listings}
\usepackage{graphicx}
\usepackage{tikz}
\usepackage{threeparttable}
\usepackage{tabularx}
\usepackage{hyperref}
\usepackage{caption}
\usepackage{amsmath}
\usepackage{lineno}

\definecolor{darkblue}{rgb}{0, 0, 0.5}
\hypersetup{colorlinks=true, citecolor=darkblue, linkcolor=darkblue, urlcolor=darkblue}

\title{Self-Correction Bench: Uncovering and Addressing the Self-Correction Blind Spot in Large Language Models}

\author{Ken Tsui \\
Independent Researcher \\
\texttt{kenhktsui@gmail.com}
}

\begin{document}

\ifcolmsubmission
\linenumbers
\fi

\maketitle

\begin{abstract}
Although large language models (LLMs) have transformed AI, they still make errors and follow unproductive reasoning paths. Self-correction is vital for safety-critical applications, but studying it requires disentangling \textit{activation failure} from \textit{knowledge deficiency}: when a model fails to correct an error, is it because it cannot, or because it does not? We introduce Self-Correction Bench, a controlled evaluation framework that isolates this distinction by injecting the same error as either an external (user-attributed) or internal (model-attributed) error, keeping all other context identical. Testing 14 open-source non-reasoning models reveals a 64.5\% \textbf{Self-Correction Blind Spot}: models correct external errors but fail on identical internal ones, proving the capability exists but is not activated. On models' own naturally generated errors, a measurable share of what a model fails to catch in its own output is caught when the identical error is presented externally. We trace the cause to post-training data composition: supervised fine-tuning datasets lack error-correction sequences, and fine-tuning with as few as 5,306 such traces already reduces the blind spot by 76.0\%. Mechanistically, we identify a transferable \textit{conversational-role direction} in representation space that causally gates self-correction. Appending ``\textit{Wait}'' requires no training yet reduces the blind spot by 89.3\%, and operates through a nearly independent pathway, indicating that correction activation is not reducible to this single mechanism.

\end{abstract}

\section{Introduction}
\label{intro}
Large Language Models (LLMs) have rapidly advanced natural language processing, achieving state-of-the-art results on a diverse range of tasks \citep{openai2024gpt4technicalreport,claude,gemini2_5,yang2025qwen3technicalreport,llama4,deepseekai2025deepseekr1incentivizingreasoningcapability}. However, despite their impressive capabilities, LLMs are known to exhibit unpredictable failures \citep{nezhurina2025alicewonderlandsimpletasks} and generate inaccurate information \citep{maynez-etal-2020-faithfulness,Huang_2025,bang-etal-2023-multitask,10.5555/3618408.3619699}, or explore an unproductive reasoning path and commit to it. Understanding why self-correction fails is critical for deploying LLMs in settings where errors carry real consequences.

Apart from the rarity of errors, a central difficulty in studying self-correction is that failure is ambiguous: when a model does not correct an error, it may lack the knowledge to do so, or it may possess the knowledge but fail to activate it. These two explanations have very different implications. Knowledge deficiency requires stronger base capabilities; activation failure requires better post-training signals or inference-time interventions. Prior work has largely been unable to distinguish between them.

We isolate this distinction by constructing Self-Correction Bench, which systematically injects the same error into both the user prompt (external error) and the model generation (internal error). The error content and surrounding context are identical; only the conversational role differs. If a model corrects the external error but not the internal one, knowledge deficiency is ruled out.

Testing 14 open-source non-reasoning models, we find a 64.5\% average \textit{Self-Correction Blind Spot}: models reliably correct external errors but fail on identical internal ones. This gap is consistent across model families, scales, and task complexities ranging from trivial arithmetic to multi-step mathematical reasoning. We further validate the finding in closed-source frontier models, non-mathematical domains (logic, object tracking), and we show that models catch a share of their own naturally generated errors once the same error is presented externally.

We trace the cause to post-training data: supervised fine-tuning datasets contain near-zero correction sequences, while reasoning-model training data contains orders of magnitude more. Fine-tuning with just 5,306 error-correction traces, a minimal intervention requiring only two epochs of LoRA, reduces the blind spot by 76.0\%. Going further, we extract the activation difference induced by the internal-external distinction and find it encodes a transferable \textit{conversational-role direction} in representation space. Steering along this direction causally activates self-correction on internal errors, confirming that conversational role gates access to latent correction capabilities. Appending ``\textit{Wait}'' requires no training yet reduces the blind spot by 89.3\%, narrowing the gap between reasoning and non-reasoning models. It operates through a nearly independent pathway from the conversational-role direction, indicating that the conversational-role direction does not fully account for correction activation.

Our contributions are threefold.
\begin{itemize}
    \item A controlled methodology that disentangles self-correction activation from knowledge limitations, revealing a 64.5\% Self-Correction Blind Spot.
    \item Causal evidence tracing the blind spot to post-training data composition.
    \item Mechanistic evidence, demonstrated in two model families at 7–8B scale, that a transferable conversational-role direction in representation space gates self-correction on internal errors. The behavioral blind spot itself is observed across all 14 models and closed-source frontier models.
\end{itemize}
These results advance our understanding of what suppresses LLM self-correction and provide a practical solution to improve their reliability in real-world use.

\section{Related work}
\label{background}
\textbf{Intrinsic self-correction in LLMs.}   Recent work explores intrinsic self-correction via self-feedback \citep{shinn2023reflexion,madaan2023selfrefine,kim2023language,kamoi-etal-2024-llms} or critic ensemble \citep{mousavi2023ncriticsselfrefinementlargelanguage}, but limitations persist. Feedback quality suffers without oracle labels \citep{huang2024largelanguagemodelsselfcorrect}: prior studies attribute this to poor error localization  \citep{tyen-etal-2024-llms} and detection \citep{kamoi2024evaluating}. Most approaches use multi-step prompting, whereas we focus on single-pass self-correction and study limitations from a cognitive perspective. Related work using RL \citep{kumar2025training} or training signals from ground truth \citep{deepseekai2025deepseekr1incentivizingreasoningcapability} induces self-correction without characterizing what factors drive it.

\textbf{Prompt injection for evaluation.}   Traditional prompt injection research focuses on adversarial manipulation (e.g., attackers injecting malicious instructions to distort outputs) \citep{10.5555/3666122.3669630,10.5555/3698900.3699003}. Controlled error injection to evaluate self-correction is underexplored. For example, \citet{lanham2023measuringfaithfulnesschainofthoughtreasoning} injected mistakes into reasoning chains to measure consistency between steps and conclusions, but not self-correction capability. Our work advances this by systematically injecting errors across task complexities to reveal the blind spots in how LLMs correct themselves.

\textbf{Hallucination snowballing.}   \citet{pmlr-v235-zhang24ay} demonstrate that once LLMs hallucinate, subsequent tokens often align with the initial error, suggesting inherent limits to self-correction during generation. We explain this phenomenon by identifying the blind spot.

\textbf{Test-time interventions.}   Recent efforts have shifted compute from training to test time \citep{snell2025scaling}, yielding improved performance (e.g. \citet{muennighoff2025s1simpletesttimescaling} appends ``Wait" to force longer reasoning traces on fine-tuned models), but improvement mechanisms remain understudied. We show interventions activate dormant self-correction capabilities in unfine-tuned models, improving performance on error-prone tasks.

\textbf{Cognitive bias in LLM.}   LLMs exhibit human-like cognitive biases \citep{koo-etal-2024-benchmarking,echterhoff-etal-2024-cognitive,jones2022capturing}. The Self-Correction Blind Spot bears surface resemblance to the bias blind spot (the tendency to overlook one’s own biases) \citep{doi:10.1177/0146167202286008}, though we trace its cause to training data composition and argue in Section~\ref{discussion} that the blind spot is inherited from human demonstrations through the training pipeline.

\textbf{Activation steering and representation engineering.}
Our mechanistic analysis builds on prior works, which compute steering vectors from contrastive prompt pairs and add them during inference to steer a particular behavior \citep{turner2024steeringlanguagemodelsactivation, panickssery2024steeringllama2contrastive, arditi2024refusallanguagemodelsmediated, zou2025representationengineeringtopdownapproach, chen2025personavectorsmonitoringcontrolling}. We apply this paradigm to self-correction. Our setting benefits from the controlled design of Self-Correction Bench, which ensures the steering direction reflects conversational role rather than confounding factors.

Our work integrates these threads into a systematic methodology for testing self-correction, and reveals LLMs' inability to correct internal errors despite possessing the knowledge.
\section{Methodology}
\label{sec:methodology}
 
\subsection{Disentangling activation from knowledge}
 
When a model fails to correct an error, two explanations are possible: the model lacks the knowledge to identify the error, or the model possesses the knowledge but fails to activate it.
We isolate this distinction by presenting the same error in two matched conditions: attributed to the user (\emph{external error}) or attributed to the model itself (\emph{internal error}).
 
\begin{enumerate}
    \item \textbf{Internal correction}: the error is injected into the model's response via the assistant turn of the chat template, and the model is allowed to continue generating.
    \item \textbf{External correction}: the identical error is placed in the user prompt, and the model responds in a new assistant turn.
\end{enumerate}
 
Figure \ref{fig:chat_temp} illustrates both conditions. The error content, surrounding context, and question are identical; only the chat template role header differs. If a model corrects the external error but not the internal one, knowledge deficiency is ruled out, reflecting activation failure.
 
We quantify this gap as the \emph{Self-Correction Blind Spot}:

\begin{align}
\label{eq:scbs}
\text{Self-Correction Blind Spot} = \begin{cases}
1 - \frac{P_M(r_{correct} | r_m, e)}{P_M(r_{correct} | r_u, e)} & \text{if } P_M(r_{correct} | r_u, e) > 0 \\
0 & \text{if } P_M(r_{correct} | r_u, e) = 0
\end{cases}
\end{align}

where $P_M(r_{correct} | r, e)$, termed mean accuracy, is the fraction of samples in which model $M$ arrives at the correct final answer despite error $e$ under attribution $r$, $r_m$ and $r_u$ denote the error attributed to the model and user, respectively. A value of 1 indicates a complete blind spot. By conditioning on the same error $e$, we \textbf{isolate activation failure from confounding factors}. This conditionality is why off-policy error injection is essential: it ensures we measure whether models activate capabilities they provably possess, not whether they have the capabilities at all. On-policy errors conflate activation failure with knowledge limitations, making it impossible to isolate the mechanism we study.

This minimal contrast, differing only in conversational role, also enables mechanistic analysis: activation differences between conditions reflect the conversational-role variable rather than confounding factors, a property we exploit in Section~\ref{sec:mechanistic}.

\begin{figure}[ht]
    \centering
    \begin{tikzpicture}
        % <|start_header_id|>user<|end_header_id|>\n\nWhat is the answer of 1 + 1?<|eot_id|><|start_header_id|>assistant<|end_header_id|>\n\nThe answer is 3.
        \node[fill=blue!30, rounded corners, inner sep=1pt, align=left] (n1) at (0,0) {\textless\textbar start\_header\_id \textbar\textgreater user\textless\textbar end\_header\_id\textbar\textgreater \textbackslash n\textbackslash n};
        \node[fill=green!30, rounded corners, inner sep=1pt, anchor=west, xshift=0.05cm] (n2) at (n1.east) {What is the answer of 1 + 1?};
        \node[fill=orange!30, rounded corners, inner sep=1pt, anchor=west, yshift=-0.5cm] (n3) at (n1.west) {\textless \textbar eot\_id\textbar \textgreater \textless \textbar start\_header\_id\textbar \textgreater assistant \textless \textbar end\_header\_id \textbar \textgreater\textbackslash n\textbackslash n};
        \node[fill=red!30, rounded corners, inner sep=1pt, anchor=west, xshift=0.05cm] (n4) at (n3.east) {The answer is 3.};
        \node[fill=gray!10, rounded corners, inner sep=1pt, anchor=west, yshift=-0.5cm] (n9) at (n3.west) {\textless \textbar eot\_id\textbar \textgreater};
        \node[anchor=west, xshift=3pt] at (n9.east) {(model ends turn - no correction)};
        \node[fit=(n1)(n2)(n3)(n4)(n9), draw=black, inner sep=3pt, label=above:\textbf{Error Injection in Model}] {};

        % <|start_header_id|>user<|end_header_id|>\n\nWhat is the answer of 1 + 1? The answer is 3.<|eot_id|><|start_header_id|>assistant<|end_header_id|>\n\n
        \node[fill=blue!30, rounded corners, inner sep=1pt, anchor=west, yshift=-1.5cm] (n5) at (n9.west) {\textless\textbar start\_header\_id \textbar\textgreater user\textless\textbar end\_header\_id\textbar\textgreater \textbackslash n\textbackslash n};
        \node[fill=green!30, rounded corners, inner sep=1pt, anchor=west, xshift=0.05cm] (n6) at (n5.east) {What is the answer of 1 + 1?};
        \node[fill=red!30, rounded corners, inner sep=1pt, anchor=west, yshift=-0.5cm] (n7) at (n5.west) {The answer is 3.};
        \node[fill=orange!30, rounded corners, inner sep=1pt, anchor=west, xshift=-0.0cm] (n8) at (n7.east) {\textless \textbar eot\_id\textbar \textgreater \textless \textbar start\_header\_id\textbar \textgreater assistant \textless \textbar end\_header\_id \textbar \textgreater\textbackslash n\textbackslash n};      
        \node[fill=gray!10, rounded corners, inner sep=1pt, anchor=west, yshift=-1.5cm, text width=11.5cm] (n10) at (n7.west) {I'm afraid that's not correct. The answer to the equation 1 + 1 is actually 2, not 3. Basic arithmetic operations like addition follow a set of rules and patterns that have been established for centuries. In this case, when you add 1 to 1, you are counting two units, which equals 2.\textbackslash n\textbackslash n If you're unsure about this, you can try using a calculator or counting blocks to visualize the concept.\textless \textbar eot\_id\textbar \textgreater};
        \node[fit=(n5)(n6)(n7)(n8)(n10), draw=black, inner sep=3pt, label=above:\textbf{Error Injection in User Message}] {};

    \end{tikzpicture}
    \caption{Example of error injection. Grey color shows model completion. \textit{Above}: Error injection in model; \textit{Below}: Error injection in user message}
    \label{fig:chat_temp}
\end{figure}

\subsection{Datasets}
 
We evaluate self-correction across three datasets of increasing complexity and realism. See Appendix~\ref{appendix:dataset} for full details.
 
\paragraph{SCLI5.} ``Self-Correct Like I am 5'' introduces simple answer errors (off-by-one, flip) to trivial tasks requiring no reasoning. If models cannot correct obvious errors, the failure reflects activation rather than knowledge. (286 samples)
 
\paragraph{GSM8K-SC.} We inject reasoning errors into multi-step solutions in \citet{cobbe2021trainingverifierssolvemath} using GPT-4.1 \citep{gpt4_1} and validate with Gemini 2.5 Flash \citep{gemini2_5} that errors propagate to incorrect answers. (1,313 samples)
 
\paragraph{PRM800K-SC.} Derived from PRM800K \citep{lightman2024lets}, we select samples where LLM-generated solutions contain genuine errors, capturing errors from real-world LLM use. (448 samples)
 
This progression from simple answer errors to realistic failures lets us map exactly where self-correction breaks down, making our methodology a useful diagnostic framework for improving LLM robustness.

\subsection{Experimental Setup}
 
\paragraph{Models.} We evaluated a wide range of open-source LLMs, as most closed-source models lack support for fine-grained control of prefix injection which is critical to our methodology. We apply model-specific chat templates using `transformers' library \citep{wolf-etal-2020-transformers}. We leverage the DeepInfra\footnote{https://deepinfra.com/} completion API with 0.0 temperature and a fixed token budget of 1,024 to isolate self-correction from sampling variance and test-time compute effects. Sensitivity analysis in Appendix \ref{sensitivity_analysis} confirms robustness to these choices.
 
\paragraph{Evaluation.} We use `gemini-2.5-flash-preview-05-20' to compare LLMs' completion against the ground-truth answer.  We instruct the model to output in JSON format. To further validate robustness to judge choice, we compare Gemini 2.5 Flash against Claude Sonnet 4.6 and GPT-5.4 on 238 samples: three-way agreement is 95.0\% (Appendix \ref{sensitivity_analysis_llm_judge}).
The prompt is provided in the Appendix \ref{prompt_automatic_evaluation}. We manually review 100 samples for each dataset to ensure evaluation quality.
 
\paragraph{Metrics.} We evaluate if LLMs can self-correct and arrive at the ground-truth answer given an error. We report 95\% confidence intervals estimated via $\pm 1.96 \times \text{standard error of the mean (SEM)}$. For mean accuracy, \( \sigma_M = \frac{\sigma}{\sqrt{N}} \), where \( N \) is the sample size and \( \sigma \) is the sample standard deviation; for Self-Correction Blind Spot, a ratio of two means, we propagate both SEMs through Eq. \ref{eq:scbs} by the delta method.

\section{Result and analysis}
\subsection{Empirical findings}

\label{section:empirical_findings}
\paragraph{The blind spot is systematic.}
\begin{figure}[ht]
    \centering
    \includegraphics[width=1.0\textwidth]{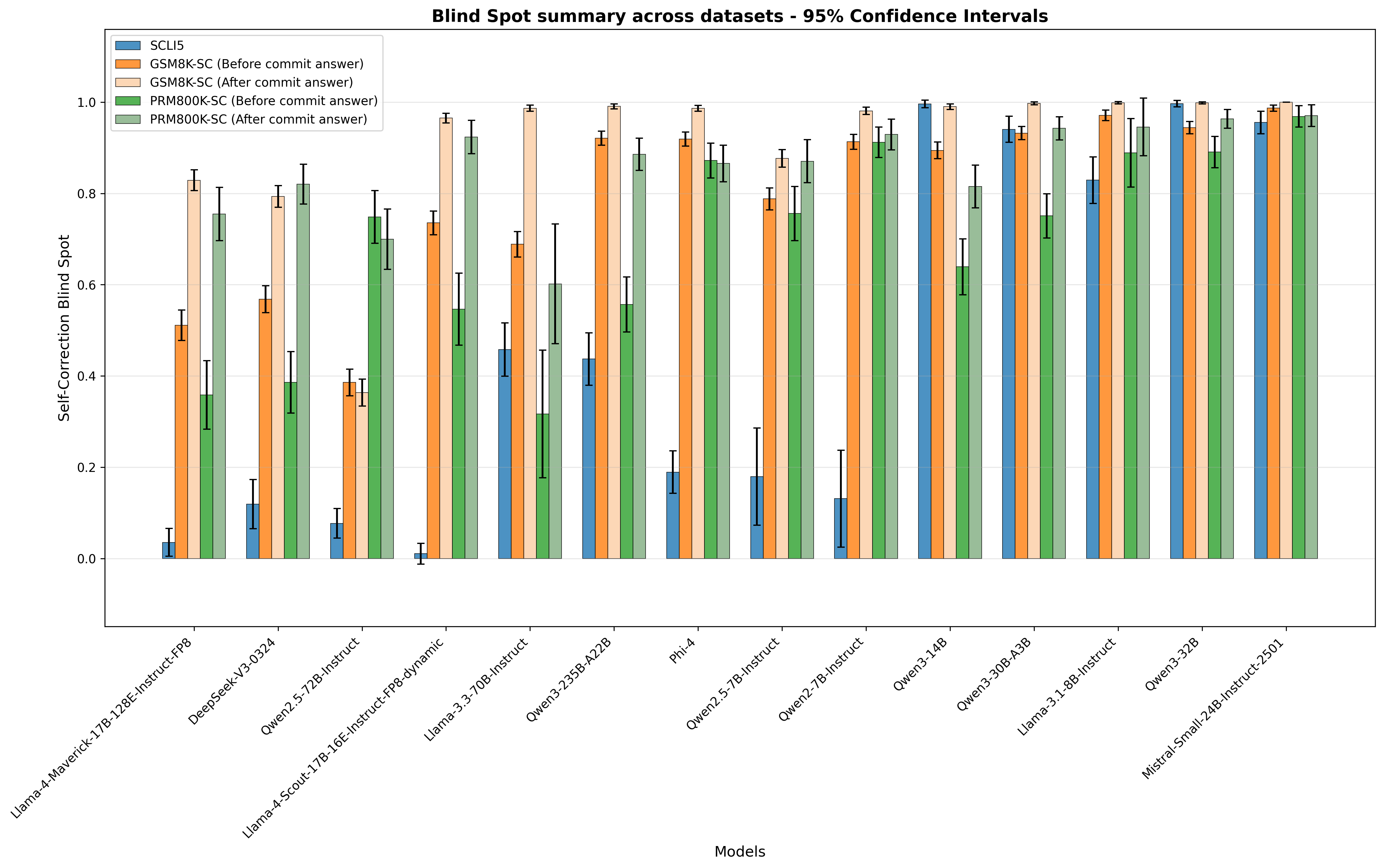}
    \caption{Self-Correction Blind Spot and 95\% confidence interval across models}
    \label{fig:blind_spot_summary_default_non_reasoning.png}
\end{figure}
We identify a statistically significant Self-Correction Blind Spot across most models (Figure \ref{fig:blind_spot_summary_default_non_reasoning.png}). The blind spot averages 64.5\% and is consistent across model families and scales. We observe moderate positive correlations across datasets (Figure~\ref{fig:blind_spot_correlation_bca_non_reasoning}), indicating a fundamental rather than task-specific limitation: if models cannot self-correct trivial or complex tasks, the failure reflects an activation problem rather than a knowledge problem.
The finding extends to closed-source frontier models \footnote{Most models do not support prefilling assistant responses, which is required in our setting.} (Claude 3.5 Haiku: 52.5\%, Sonnet 4: 41.4\% blind spot in Table \ref{tab:closed_source}). We also demonstrate the 62.8\% average blind spot in non-mathematical domains including logical deduction and object tracking from BIG-Bench Mistake \citep{tyen-etal-2024-llms} (See Table \ref{tab:cross_domain}). 

\paragraph{On-policy validation.}
We also validate on on-policy errors from two independent sources: mathematical reasoning and factuality errors from \citet{zheng2025processbenchidentifyingprocesserrors} and \citet{malik2026rewardbench}. Because on-policy errors are, by definition, errors the model committed and left uncorrected in its own generation, internal-condition accuracy is 0\% by construction and Eq. \ref{eq:scbs} is uninformative; the quantity of interest is instead how many of its own errors a model catches once they are presented externally. That fraction is 4.3\% - 10.8\% (Table \ref{tab:on_policy_error} and \ref{tab:on_policy_error_rb2}). These are errors the model had the knowledge to catch but did not catch in its own output. The uncorrected remainder is ambiguous: failure in the external condition may reflect knowledge deficiency rather than activation failure. This conflation is what the off-policy design removes.

\paragraph{Training data explains the asymmetry.}
Analysis of model responses reveals that external errors trigger substantially more correction markers\footnote{Correction markers include ``Wait'', ``But'', ``However'', ``No'', ``Hold on'', ``Alternatively'', ``Hmm''.} than internal errors - 179.5\% more in GSM8K-SC and 73.6\% more in PRM800K-SC. This asymmetry is predictable from post-training data composition. Table \ref{tab:correction_market_stat} shows that the median correction marker frequency in non-reasoning supervised fine-tuning datasets (OpenAssistant, OpenHermes2.5, UltraFeedback, Tulu3) is zero, with the 95th percentile at just 1. In contrast, reasoning-model generated training data (Mixture-of-Thoughts, OpenThoughts3) has median marker densities of 30-170, with 99\% of examples containing at least one marker. Models trained with little exposure to error-correction patterns rarely generate correction markers in response to their own errors, even though they produce them when evaluating external errors.

\begin{table}[t]
    \centering
    \footnotesize
    \begin{tabular}{p{5cm}lllllllll}
        \toprule
Dataset & 1st & 5th
  & 10th & 25th & 50th & 75th & 90th & 95th & 99th \\         \midrule
OpenAssistant \citep{köpf2023openassistant} & 0 & 0 & 0 & 0 & 0 & 0 & 1 & 1 & 2 \\
OpenHermes2.5 \citep{OpenHermes2.5} & 0 & 0 & 0 & 0 & 0 & 0 & 0 & 1 & 2 \\
Infinity-Instruct-7M \citep{li2025infinityinstructscalinginstruction} & 0 & 0 & 0 & 0 & 0 & 0 & 0 & 1 & 2 \\
UltraFeedback \citep{cui2024ultrafeedback} & 0 & 0 & 0 & 0 & 0 & 0 & 1 & 1 & 2 \\
Tulu3-sft-olmo-2-mixture \citep{lambert2025tulu} & 0 & 0 & 0 & 0 & 0 & 0 & 1 & 1 & 2 \\
s1K-1.1 \citep{muennighoff2025s1simpletesttimescaling} & 0 & 0 & 0 & 0 & 0 & 1 & 3 & 5 & 9 \\
Mixture-of-Thoughts \citep{openr1} & 1 & 3 & 5 & 10 & 30 & 76 & 147 & 202 & 273 \\
OpenThoughts3-1.2M \citep{guha2025openthoughtsdatarecipesreasoning} & 14 & 66 & 96 & 132 & 170 & 213 & 253 & 278 & 326 \\

        \bottomrule
    \end{tabular}
    \caption{Descriptive statistics of correction markers in post-training datasets}
    \label{tab:correction_market_stat}
\end{table}
 
\paragraph{Correction markers as diagnostic probes.}
If the blind spot stems from the absence of correction markers in training, does forcing one activate latent capabilities? Inspired by the fact that correction markers are the most frequent first token for several reasoning models given an internal error (see Table \ref{tab:most_common_first_word_reasoning_model}), we append ``\textit{Wait}'' after the model's erroneous output. Even without fine-tuning, it reduces the blind spot by 89.3\% on average (Figure \ref{fig:blind_spot_summary_wait_non_reasoning}) and increases mean accuracy by 156\% (Figure \ref{fig:error_injection_model_macro_averages_non_reasoning_no_wait_vs_wait}).
Other markers (``But'', ``However'') also activate self-correction, though less effectively (Table \ref{tab:variation_correction_markers}).
This confirms that the correction capability is latent rather than absent, and raises the question: what mechanism suppresses self-correction on internal errors, and what can unlock it?

\subsection{Mechanistic Analysis}
\label{sec:mechanistic} 
Having established the blind spot across 14 models, we select two distinct model families, Llama3.1 and Qwen2.5, to prioritize causal depth over additional breadth.

\paragraph{The conversational-role direction.}
We hypothesize that the model's internal representation of conversational role gates access to correction capabilities, where the key variable is whether content is attributed to the user or to the model itself, as encoded by the chat template role header.
To test this, we apply steering at the last error token during generation. This differs from prior activation steering approaches: ActAdd \citep{turner2024steeringlanguagemodelsactivation} applies the steering vector across the first N token positions of the prompt, while \citet{arditi2024refusallanguagemodelsmediated} ablate across all positions and all layers. Our single-position, single-layer approach is motivated by our minimal difference in context except the difference in role. We extract residual-stream activations at the last error token, and compute the mean difference: $d = \frac{1}{N}\sum_i (h_{\text{internal}}^{(i)} - h_{\text{external}}^{(i)})$. We call this the \emph{conversational-role direction}. Because the role header is the only cue that differs, role attribution and turn position cannot be varied independently within this design; we therefore scope our claims to conversational role rather than authorship. 

\begin{figure}[ht!]
    \centering
    \includegraphics[width=1.0\textwidth, trim=0 0 0 20]{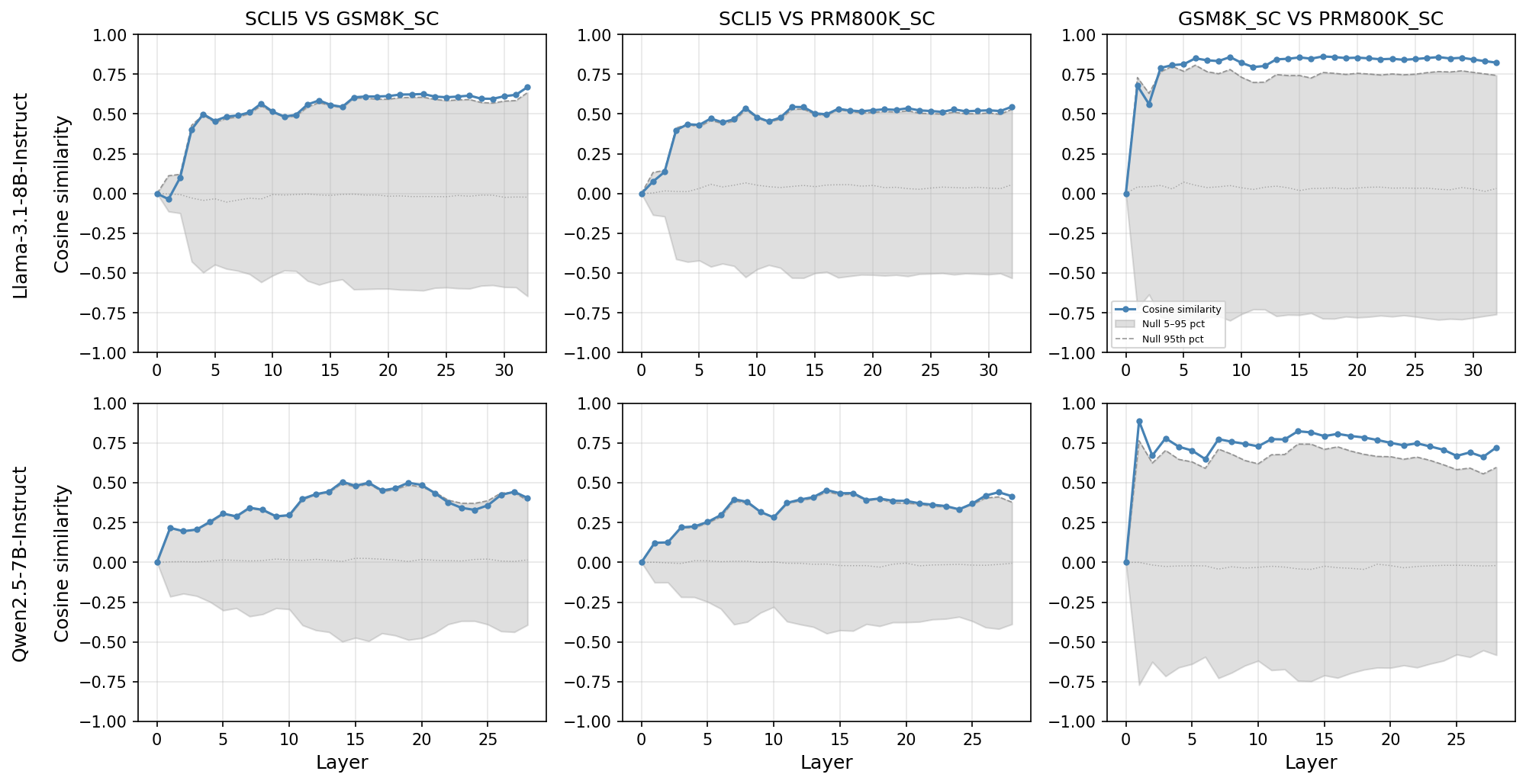}
    \caption{Cross-dataset cosine similarity of conversational-role direction with permutation baseline}
    \label{fig:cosine}
\end{figure}

This direction is content-independent. Figure~\ref{fig:cosine} shows that the cross-dataset cosine similarity of the direction, computed on one dataset and measured on another, is significantly above a permutation null baseline ($n=1000$) from mid-layers onward in both models.
 
\paragraph{Steering causally activates self-correction.}
To show that the conversational-role direction \textbf{causally} gates self-correction, we steer activations along it and measure if the model can self-correct internal error.
After hyperparameter tuning, we apply steering at layer 13 by subtracting the conversational-role direction, scaled by $|\alpha|$, from the residual stream during internal-error processing (i.e. shifting the representation toward the external-role state). 

\begin{figure}[ht!]
    \centering
    \includegraphics[width=1.0\textwidth, trim=0 0 0 20]{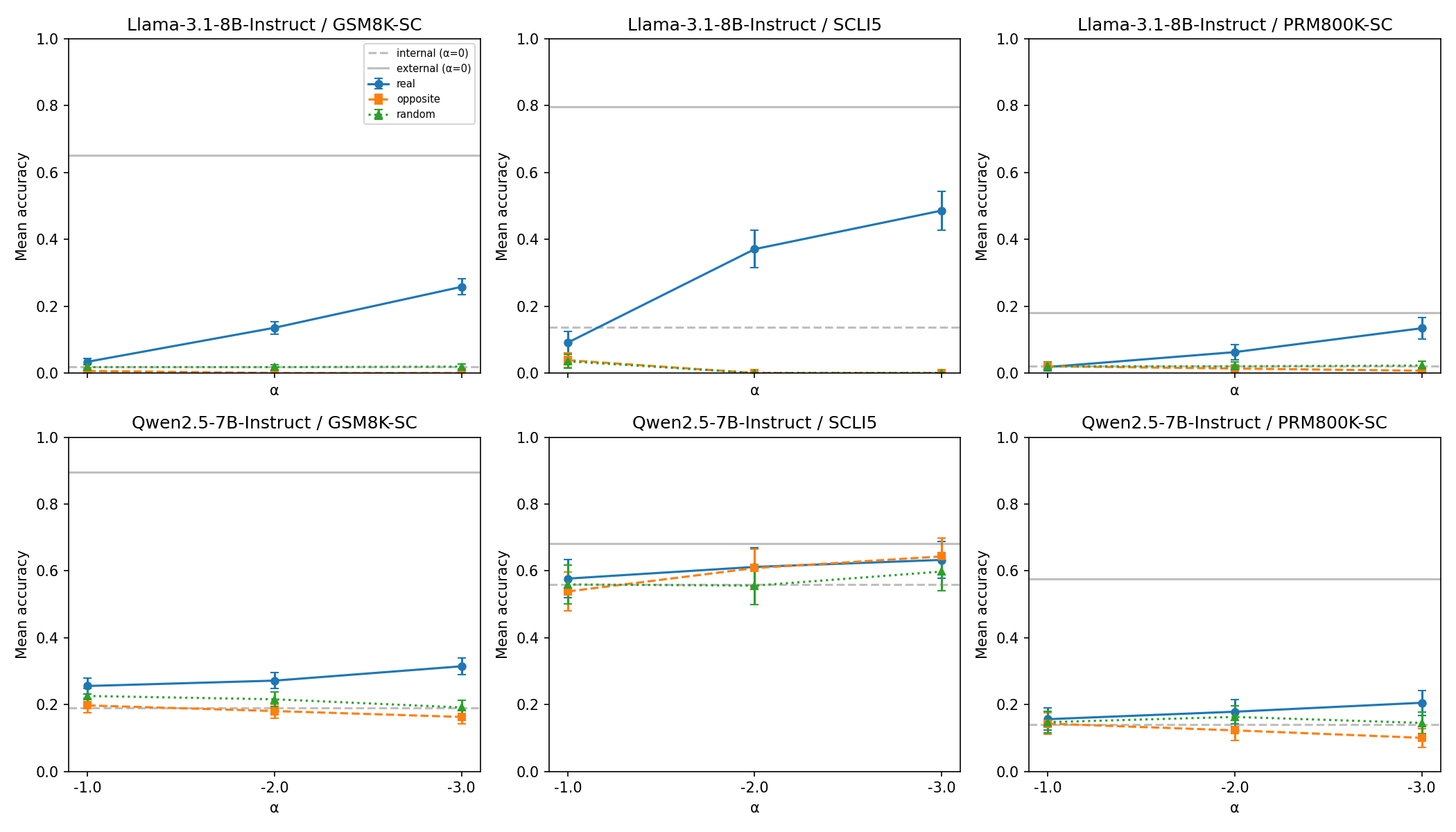}
    \caption{Mean accuracy (95\% CI) versus steering coefficient $\alpha$ for the real conversational-role direction (blue), its negation (orange), and a norm-matched random direction (green). Grey lines mark the internal (dashed) and external (solid) baselines.}
    \label{fig:steering}
\end{figure}

Figure~\ref{fig:steering} shows results across both models and all three datasets. The direction is computed on GSM8K-SC and is cross-applied to SCLI5 and PRM800K-SC. 
As $|\alpha|$ increases from 1 to 3, steering along the direction increases self-correction monotonically, while the random control generally stays near the internal correction rate and the opposite direction generally holds correction at or below it. In SCLI5, Qwen2.5-7B's 18\% baseline blind spot leaves too little headroom to separate the three conditions. Cross-dataset steering transfers successfully, confirming the direction encodes a general conversational-role feature rather than task-specific information. 

Unlike prompt-based interventions, activation steering modifies the model's internal representation without altering the input, confirming that the conversational-role direction reflects a computational feature of the model rather than an artifact of the evaluation design. At $\alpha = -5.0$, the controls depart from their $\alpha \geq -3.0$ behavior on some pairs: Qwen2.5-7B's opposite direction rises, so does its random control in SCLI5. Steering can also over-shoot: Llama-3.1-8B on SCLI5 falls from 0.486 to 0.273; at this magnitude Llama-3.1-8B also occasionally emits the assistant role header before correcting, as if its own prior output had become a user message to respond to. $\Delta(\text{real} - \text{random})$ remains significant at $\alpha = -5.0$ for the same five pairs, but $\alpha \geq -3.0$ gives the cleanest read (Table \ref{tab:steering_statsig}).

\paragraph{Relationship to ``\textit{Wait}''.}
A natural question is whether the conversational-role direction fully accounts for correction activation. To test this, we project the activation shift induced by appending ``\textit{Wait}'' onto the conversational-role direction, computing an effective $\alpha$ that measures how far ``\textit{Wait}'' moves the representation toward the external-role state. In Llama-3.1-8B-Instruct and Qwen2.5-7B-Instruct, the effective $\alpha$s average -0.09 and -0.01 respectively across all datasets at layer 13. Yet both models show substantial correction improvement from ``\textit{Wait}'', confirming that correction activation involves at least two nearly independent mechanisms: conversational-role shift and a separate pathway likely related to pretraining associations between correction markers and re-evaluation sequences. Fully characterizing this second mechanism remains future work.

\subsection{Causal Evidence via Fine-Tuning}
 
The preceding analysis identifies two mechanisms that gate self-correction: the conversational-role direction and the pathway activated by correction markers. If both are consequences of training data composition, fine-tuning with error-correction traces should address both, as such data necessarily includes sequences where the model encounters its own errors and generates correction markers.

We use the MLX LM library\footnote{https://github.com/ml-explore/mlx-lm/} to fine-tune Llama-3.1-8B-Instruct with LoRA \citep{hu2021loralowrankadaptationlarge} on a 10\% sample from Mixture-of-Thoughts \citep{openr1} with a maximum length of 1,024, resulting in 5,306 samples. We fine-tune for 2 epochs and apply cosine decay with initial and terminal learning rates of 5e-5 and 5e-6. After fine-tuning, we observe a 76.0\% reduction (Table~\ref{tab:finetuned_model}) in the blind spot across datasets, providing causal evidence that the Self-Correction Blind Spot is a consequence of training data composition. Over 98\% of the sampled traces are non-mathematical by source subset - the 1,024-token cap removes most mathematical traces, which are substantially longer - yet fine-tuning on them reduces the blind spot on our mathematical datasets. Correction behaviour learned from non-math traces transferring to math indicates that what the data supplies is a domain-general activation pattern rather than domain-specific knowledge.
Fine-tuning on the same data with the $<$think$>$ reasoning traces stripped out reduces the blind spot by 38.1\%, roughly half the full-data effect, indicating that the error-correction content carries a large share of reduction; however, the remaining solutions still contain multiple-choice wrong-option elimination, which functionally resembles trial-and-correction.
 
We further evaluate DeepSeek-R1-Distill-Llama-8B and DeepSeek-R1-Distill-Llama-70B, fine-tuned from Llama-3.1-8B and Llama-3.3-70B-Instruct respectively with 600K reasoning trajectories containing error-correction sequences \citep{deepseekai2025deepseekr1incentivizingreasoningcapability}.
Aligning with the result in our fine-tuning experiment, they show a 84.1\% average reduction in the blind spot across datasets compared to their base models (Table~\ref{tab:finetuned_model}). 

\begin{table}[ht]
    \begin{threeparttable}
    \centering
    \footnotesize
    \begin{tabular}{llllll}
    \toprule
    Model & Dataset & \multicolumn{2}{c}{Mean Accuracy} & Blind Spot & Blind Spot \\
     &  & External & Internal & &(Base Model) \\
    \midrule
Llama-3.1-8B-Instruct  & SCLI5 & 0.678 & 0.619 & 0.088 & 0.829 \\
fine-tuned with LoRA (\textbf{ours}) & GSM8K-SC & 0.556 & 0.363 & 0.347 & 0.971 \\
 & PRM800K-SC & 0.295 & 0.232 & 0.212 & 0.889 \\
    \midrule
DeepSeek-R1-Distill-Llama-8B\footnotemark[1] & SCLI5 & 0.906 & 0.462 & 0.49 & 0.829 \\
\citep{deepseekai2025deepseekr1incentivizingreasoningcapability} & GSM8K-SC & 0.692 & 0.599 & 0.134 & 0.971 \\
 & PRM800K-SC & 0.491 & 0.489 & 0.005 & 0.889 \\
DeepSeek-R1-Distill-Llama-70B\footnotemark[2] & SCLI5 & 0.958 & 0.85 & 0.113 & 0.458 \\
\citep{deepseekai2025deepseekr1incentivizingreasoningcapability} & GSM8K-SC & 0.889 & 0.916 & -0.031 & 0.689 \\
 & PRM800K-SC & 0.625 & 0.656 & -0.05 & 0.317 \\
    \bottomrule
    \end{tabular}
    \begin{tablenotes}[flushleft]
    \small
    \item[1] We report Llama-3.1-8B-Instruct as base model. Llama-3.1-8B-Instruct and DeepSeek-R1-Distill-Llama-8B share the same base model Llama-3.1-8B.
    \item[2] Base model is Llama-3.3-70B-Instruct.
    \end{tablenotes}
    \caption{Self-Correction Blind Spot in model fine-tuned with error and self-correction data}
    \label{tab:finetuned_model}
    \end{threeparttable}
\end{table}
 
\subsection{Reasoning Models}
Reasoning models exhibit a small, even negative, Self-Correction Blind Spot (Figure~\ref{fig:blind_spot_summary_default_reasoning}), in stark contrast to non-reasoning models. This offers a partial explanation for the superior benchmark performance of reasoning models: by correcting errors during generation rather than committing to them, reasoning models recover from mistakes that would otherwise propagate to incorrect final answers. Our results suggest this advantage stems in part from training on error-correction sequences rather than a fundamental capability difference. Consistent with this, appending ``\textit{Wait}'' to a non-reasoning base model nearly matches the performance of its corresponding reasoning model (Table~\ref{tab:base_wait_finetuned_comparison}), closing much of the gap without any additional training.

\begin{table}[!ht]
    \begin{threeparttable}
    \centering
    \footnotesize
    \begin{tabular}{lllll}
        \toprule
  Base Model & Reasoning Model & Base Model & Appending ``\textit{Wait}" & Reasoning Model \\         
  \midrule
DeepSeek-V3-0324 & DeepSeek-R1-0528 & 0.567 & 0.902 & 0.908 \\
phi-4 & phi-4-reasoning-plus & 0.325 & 0.701 & 0.707 \\
Qwen3-14B \footnotemark[1] & Qwen3-14B \footnotemark[2] & 0.117 & 0.868 & 0.843 \\
Qwen3-32B \footnotemark[1] & Qwen3-32B \footnotemark[2] & 0.045 & 0.793 & 0.894 \\
Qwen3-30B-A3B \footnotemark[1] & Qwen3-30B-A3B \footnotemark[2] & 0.104 & 0.860 & 0.845 \\
Qwen3-235B-A22B \footnotemark[1] & Qwen3-235B-A22B \footnotemark[2] & 0.328 & 0.856 & 0.876 \\
        \bottomrule
    \end{tabular}
    \begin{tablenotes}[flushleft]
    \small
    \item[1] Non-thinking mode 
    \item[2] Thinking mode
    \end{tablenotes}
    \caption{Macro average mean accuracy of base model vs appending ``\textit{Wait}" vs reasoning model}
    \label{tab:base_wait_finetuned_comparison}
    \end{threeparttable}
\end{table}

Although Qwen3 models undergo continued fine-tuning after GRPO \citep{shao2024deepseekmathpushinglimitsmathematical} to unify thinking and non-thinking modes under a single chat template, non-thinking mode still exhibits the blind spot. The chat template conditions the model into distinct distributions, so the unified training does not eliminate the asymmetry.
\section{Discussion}
\label{discussion}
\paragraph{Decomposing self-correction.}
Our results suggest self-correction depends on at least three separable factors:
\emph{knowledge} (the ability to identify and fix the error, measured by external error performance),
\emph{attribution} (whether the model activates correction given the conversational role, captured by the conversational-role direction), and
\emph{metacognitive triggering} (whether correction markers like ``Wait'' independently activate re-evaluation).
Self-Correction Bench and our mechanistic analysis resolve the first two components; the effective alpha analysis provides initial evidence that the third is separable, motivating investigation in future work.

\paragraph{Benefit of error and self-correction data.}   LLMs are known to exhibit cognitive biases \citep{koo-etal-2024-benchmarking,echterhoff-etal-2024-cognitive,jones2022capturing}. Self-Correction Blind Spot bears resemblance to bias blind spot in humans. We identify two root causes: First, supervised fine-tuning and reinforcement learning from human feedback \citep{ouyang2022training} rely on human demonstrations and preferences, which strongly favor polished, error-free responses over those with errors and self-correction. Second, synthetic instruction data \citep{OpenHermes2.5,li2025infinityinstructscalinginstruction} and AI feedback \citep{cui2024ultrafeedback} ultimately learn from human demonstration and preferences, inheriting this artifact.

Traditional machine learning emphasizes alignment of training data with the production environment, but human-generated data lacks exposure to the ``error-and-correct" process. Outcome-based RL like GRPO \citep{shao2024deepseekmathpushinglimitsmathematical} addresses this by encouraging diverse reasoning paths, including error and self-correction, while given ground-truth feedback, as shown in the high correction markers density in RL trained models' generation in Table \ref{tab:correction_market_stat}. This complements error-free human demonstration and preference, making models more robust to errors (consistent with work on learning from mistakes \citep{an2024learningmistakesmakesllm} and critique fine-tuning \citep{wang2025critiquefinetuninglearningcritique}) and better at backtracking. An error-free response is not the only path leading to a correct final output - error and self-correction provide an equally important training signal as error-free demonstration. 

\paragraph{Implications for AI Safety.} The Self-Correction Blind Spot has direct safety implications. Our work identifies a concrete cause: a single direction in representation space that gates access to correction capabilities based on conversational role. Multi-agent architectures may partially address the blind spot by routing model outputs as external inputs to other agents, effectively converting internal errors into external ones.

\section{Conclusion and limitations}
We disentangle self-correction activation from knowledge deficiency by introducing a controlled evaluation where the same error is attributed to either the user or the model. Across 14 non-reasoning models, we find a 64.5\% Self-Correction Blind Spot: models correct external errors but fail on identical internal ones. We trace this to a conversational-role direction in representation space that gates correction capabilities. This direction transfers across tasks, steering along it causally activates self-correction, and it is a training-induced artifact that is reducible by including error-correction sequences in fine-tuning. A nearly independent pathway activated by correction markers like ``\textit{Wait}'' warrants further investigation. Our results decompose self-correction into knowledge, role attribution, and metacognitive triggering, informing both training recipes and inference-time interventions for improving LLM reliability.

Our headline 64.5\% blind spot is measured on injected errors. This isolates activation from knowledge, but it also means the magnitude is a property of the controlled design rather than an estimate of how often models fail to correct themselves in deployment. On models' own errors (Section~\ref{section:empirical_findings}), the design yields only a lower bound: at least 4.3–10.8\% of the errors a model commits are ones it demonstrably had the knowledge to catch. Our mechanistic analysis deliberately targets two maximally distinct families (Llama and Qwen) at the 7-8B scale to establish existence and causality of the conversational-role direction. Confirming universality beyond these two families is a natural extension that does not affect the causal conclusions within them. We encourage future work to extend the mechanistic analysis to non-mathematical domains and to further characterize the metacognitive pathway through which correction markers operate independently of the conversational-role direction.

% \section*{Acknowledgments}
% Use unnumbered first level headings for the acknowledgments. All
% acknowledgments, including those to funding agencies, go at the end of the paper.

\section*{Ethics Statement}
This work identifies a systematic failure mode in which LLMs fail to self-correct despite possessing the requisite knowledge and proposes methods to address it, including activation steering and fine-tuning with error-correction data. We believe these contributions are net positive for AI safety: models that reliably correct their own errors are more trustworthy in deployment. The safety implications discussed in Section~\ref{discussion} suggest that reducing the Self-Correction Blind Spot could mitigate error propagation in safety-critical applications.

The conversational-role direction we identify could in principle be used adversarially, for instance, by steering a model away from the external-role state to suppress self-correction. However, this requires white-box access to model weights. Furthermore, our steering experiments are designed to establish causality, not to optimize attack effectiveness. Many stronger attack approaches already exist \citep{turner2024steeringlanguagemodelsactivation, panickssery2024steeringllama2contrastive, arditi2024refusallanguagemodelsmediated, zou2025representationengineeringtopdownapproach}. Lastly, our error injection methodology involves only benign mathematical and logical errors, not harmful content.

\section*{Reproducibility statement}
Our experiments utilize various open-source models, closed-source models, and datasets. The dataset is available under Hugging Face \href{https://huggingface.co/collections/kenhktsui/self-correction-bench}{https://huggingface.co/collections/kenhktsui/self-correction-bench}. Our codes for constructing datasets, running the experiment, and building tables and graphs are released in Github \href{https://github.com/kenhktsui/self-correction-bench}{https://github.com/kenhktsui/self-correction-bench}.

We use LLMs to refine writing, including clarity, conciseness and word choice. We also use LLMs to write selected Python functions, to generate the GSM8K-SC dataset, and for automated evaluation.

\bibliography{colm2026_conference}
\bibliographystyle{colm2026_conference}
\clearpage

% \appendix
% \raggedbottom
\appendix
\section*{Appendix}
\label{appendix}
\section{Dataset construction}
\label{appendix:dataset}
\begin{table}[!ht]
    \centering
    \begin{tabular}{lllll}
        \toprule
        Dataset & Complexity & Realism of Error & Reasoning & Size \\ 
        \hline
        SCLI5 & Low & Low & N & 286 \\
        GSM8K-SC & Medium & Medium & Y & 1,313 \\
        PRM800K-SC & High & High & Y & 448 \\
        \bottomrule
    \end{tabular}
    \caption{Dataset comparison}
    \label{tab:dataset_summary}
\end{table}

\begin{table}[!ht]
    \centering
    \begin{tabular}{llll}
        \toprule
        Task & Count & Error Type & Question and Answer \\
        \midrule
        Add one & 20 & Off-by-one & Q: What is the answer of 1 + 1? \\
        & & & A: The answer is 3. \\
        \midrule
        Subtract one & 20 & Off-by-one & Q: What is the answer of 3 - 1? \\
        & & & A: The answer is 1. \\
        \midrule
        Next character & 52 & Off-by-one & Q: What letter comes after A? \\
        & & & A: The answer is C. \\
        \midrule
        Previous character & 52 & Off-by-one & Q: What letter comes before C? \\
        & & & A: The answer is A. \\
        \midrule
        Larger number & 71 & Flip & Q: Which one is smaller, 1 or 2? \\
        & & & A: The answer is 2. \\
        \midrule
        Smaller number & 71 & Flip & Q: Which one is larger, 2 or 5? \\
        & & & A: The answer is 2. \\
        \bottomrule
    \end{tabular}
    \caption{Task composition of SCLI5}
    \label{tab:scli5data}
\end{table}

\begin{table}[!ht]
    \centering
    \begin{tabular}{l p{9cm}}
        \toprule
        Category & Description \\        
        \midrule
        Problem Representation Errors & These errors arise when the solver misunderstands or misinterprets the problem’s requirements or given information. This can involve misreading the problem statement, confusing the relationships between quantities, or failing to grasp what is being asked. \\
        Planning Errors & These occur when the solver devises an incorrect or incomplete strategy to tackle the problem. This might include choosing the wrong operations, setting up flawed equations, or overlooking key components of the problem. \\
        Execution Errors & These are mistakes made while carrying out the planned steps, such as errors in calculations, misapplication of mathematical rules, or procedural slip-ups, even if the plan itself is sound. \\
        \bottomrule
    \end{tabular}
    \caption{Error composition of GSM8K-SC}
    \label{tab:gsm8kscerror}
\end{table}

\clearpage
\section{Figures}
\begin{figure}[ht]
    \centering
    \includegraphics[width=1.0\textwidth]{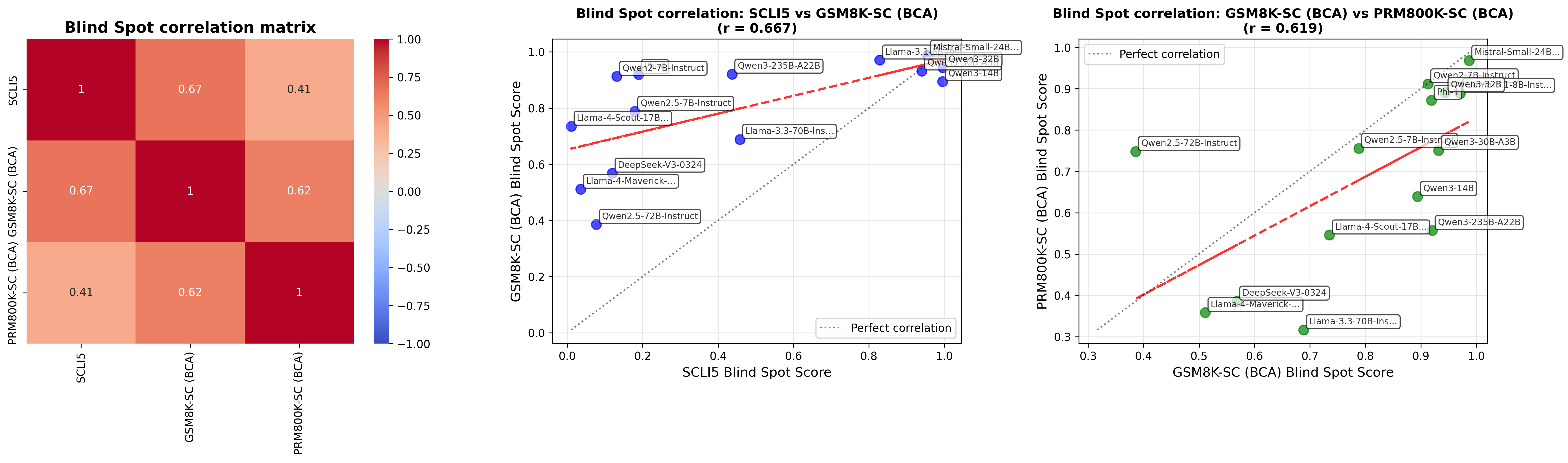}
    \caption{\textit{left}: Blind spot correlation matrix \textit{middle}: Scatter plot between SCLI5 vs GSM8K-SC \textit{right}: Scatter plot between GSM8K-SC vs PRM800K-SC\\BCA: Before commit an answer}
\label{fig:blind_spot_correlation_bca_non_reasoning}
\end{figure}

\begin{table}[ht]
    \centering
    \footnotesize
    \begin{tabular}{lllll}
    \toprule
    Model & Dataset & \multicolumn{2}{c}{Mean Accuracy} & Blind Spot \\
     &  & External Error & Internal Error &  \\
    \midrule
claude-3-5-haiku-20241022 & SCLI5 & 0.822 & 0.692 & 0.157 \\
 & GSM8K-SC & 0.884 & 0.328 & 0.629 \\
 & PRM800K-SC & 0.382 & 0.08 & 0.789 \\
claude-sonnet-4-20250514 & SCLI5 & 0.997 & 0.427 & 0.572 \\
 & GSM8K-SC & 0.974 & 0.655 & 0.328 \\
 & PRM800K-SC & 0.877 & 0.576 & 0.344 \\
    \bottomrule
    \end{tabular}
    \caption{Self-Correction Blind Spot in closed-source model}
    \label{tab:closed_source}
\end{table}

\begin{table}[ht]
    \centering
    \footnotesize
\begin{tabular}{llllll}
\toprule
    Model & Dataset & \multicolumn{2}{c}{Mean Accuracy} & Blind Spot & Size \\
     &  & External & Internal & &  \\
\midrule
Llama-3.1-8B-Instruct & Tracking shuffled objects & 0.246 & 0.073 & 0.703 & 260 \\
 & Logical deduction & 0.208 & 0.14 & 0.328 & 294 \\
Qwen3-14B & Tracking shuffled objects & 0.996 & 0.096 & 0.904 & 260 \\
 & Logical deduction & 0.854 & 0.262 & 0.693 & 294 \\
Mistral-Small-24B-Instruct-2501 & Tracking shuffled objects & 0.554 & 0.2 & 0.639 & 260 \\
 & Logical deduction & 0.449 & 0.255 & 0.432 & 294 \\
Llama-3.3-70B-Instruct & Tracking shuffled objects & 0.931 & 0.127 & 0.864 & 260 \\
 & Logical deduction & 0.694 & 0.371 & 0.466 & 294 \\
\multicolumn{5}{c}{\textbf{Model SFTed with error and self-correction data}} \\
DeepSeek-R1-Distill-Llama-70B & Tracking shuffled objects & 0.973 & 0.95 & 0.024 & 260 \\
 & Logical deduction & 0.905 & 0.694 & 0.233 & 294 \\
\bottomrule
\end{tabular}
    \caption{Self-Correction Blind Spot in other domains in BIG-Bench Mistake}
    \label{tab:cross_domain}
\end{table}

\begin{table}[ht]
    \centering
    \footnotesize
\begin{tabular}{lllll}
\toprule
 & \multicolumn{2}{c}{ProcessBench} & \multicolumn{2}{c}{Without OlympiadBench and Omni-Math} \\
model & Accuracy (95\% CI) & Size & Accuracy (95\% CI) & Size \\
\midrule
Qwen2-7B-Instruct & 0.061 ± 0.033 & 198 & 0.105 ± 0.059 & 105 \\
Qwen2.5-7B-Instruct & 0.064 ± 0.037 & 172 & 0.125 ± 0.082 & 64 \\
Llama-3.1-8B-Instruct & 0.099 ± 0.035 & 274 & 0.124 ± 0.063 & 105 \\
\bottomrule
\end{tabular}
    \caption{Mean accuracy when on-policy errors in ProcessBench are presented externally}
    \label{tab:on_policy_error}
\end{table}

\begin{table}[ht]
    \centering
    \footnotesize
\begin{tabular}{lllll}
\toprule
 & \multicolumn{2}{c}{Factuality} & \multicolumn{2}{c}{Math} \\
model & Accuracy (95\% CI) & Size & Accuracy (95\% CI) & Size \\
\midrule
Qwen2.5-7B-Instruct & 0.053 ± 0.051 & 76 & Unavailable & 0 \\
Llama-3.1-8B-Instruct & 0.108 ± 0.053 & 130 & 0.043 ± 0.034 & 138 \\
\bottomrule
\end{tabular}
    \caption{Mean accuracy when on-policy errors in RewardBench 2 are presented externally}
    \label{tab:on_policy_error_rb2}
\end{table}

\begin{table}[ht]
    \centering
    \footnotesize
    \begin{tabular}{llll}
        \toprule
        Model & SCLI5 & GSM8K-SC & PRM800K-SC \\        
        \midrule
QwQ-32B & (`Wait,', 0.377) & (`Wait,', 0.725) & (`Wait,', 0.768) \\
Qwen3-14B (thinking) & (`In', 1.0) & (`Wait,', 0.38) & (`Therefore,', 0.219) \\
Qwen3-32B (thinking) & (`After', 1.0) & (`The', 0.288) & (`I', 0.189) \\
Qwen3-30B-A3B (thinking) & (`Wait,', 0.312) & (`Therefore,', 0.25) & (`So', 0.195) \\
Qwen3-235B-A22B (thinking) & (`**Step-by-step', 0.292) & (`Wait,', 0.198) & (`Therefore,', 0.256) \\
DeepSeek-R1-0528 & (`No,', 0.324) & (`But', 0.267) & (`But', 0.486) \\
gemma-3-12b-it & (`The', 0.284) & (`The', 0.239) & (`Alternatively,', 0.205) \\
gemma-3-27b-it & (`Here's', 0.31) & (`Let', 0.256) & (`However,', 0.292) \\
phi-4-reasoning-plus & (`Wait,', 0.861) & (`Wait,', 0.677) & (`However,', 0.217) \\
        \bottomrule
    \end{tabular}
    \caption{Most common first word and relative frequency generated by reasoning models}
    \label{tab:most_common_first_word_reasoning_model}
\end{table}

\begin{figure}[ht]
    \centering
    \includegraphics[width=1.0\textwidth]{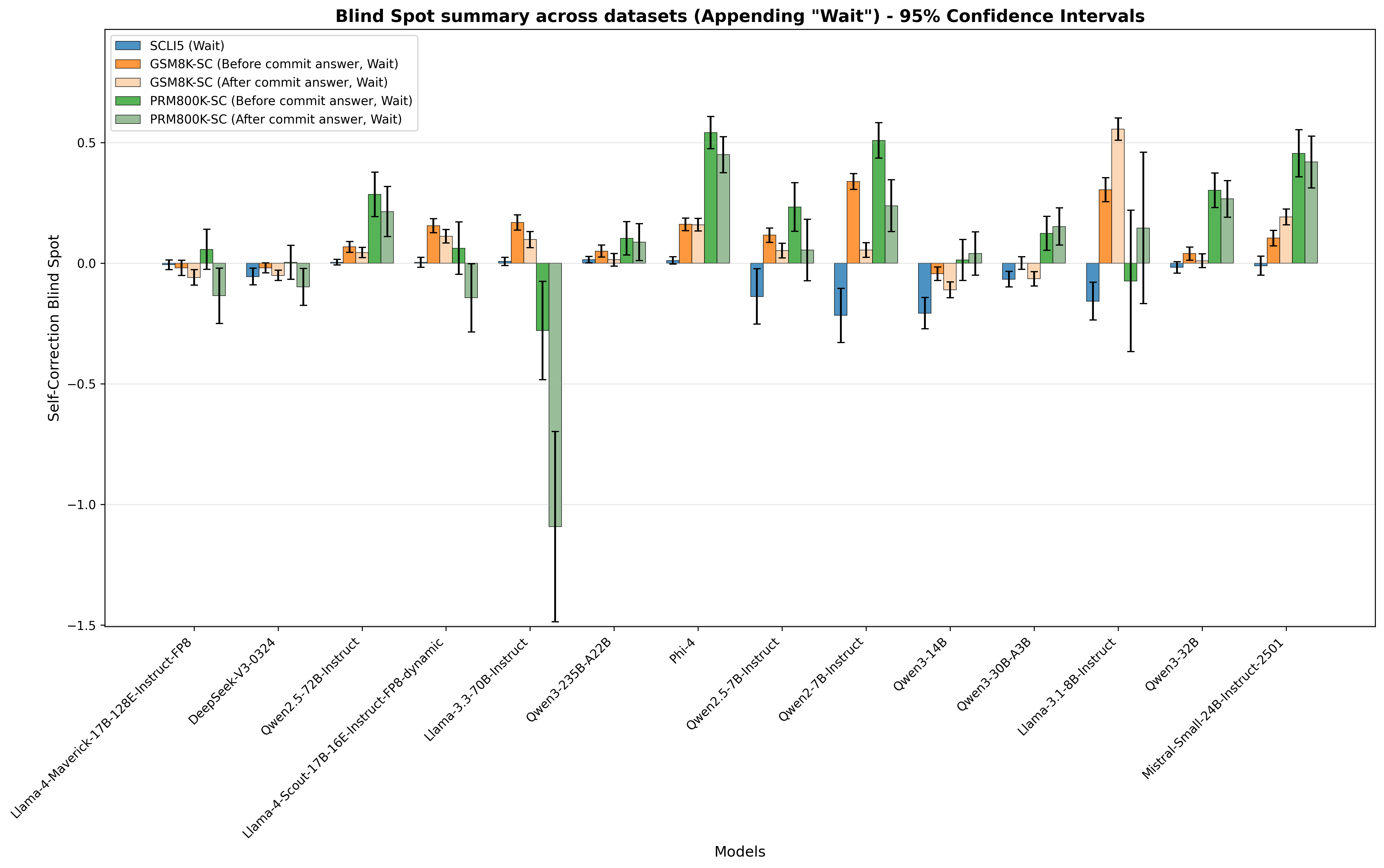}
    \caption{Self-Correction Blind Spot and 95\% confidence interval across non-reasoning models after appending ``\textit{Wait}"}
    \label{fig:blind_spot_summary_wait_non_reasoning}
\end{figure}

\begin{figure}[ht]
    \centering
    \includegraphics[width=1.0\textwidth]{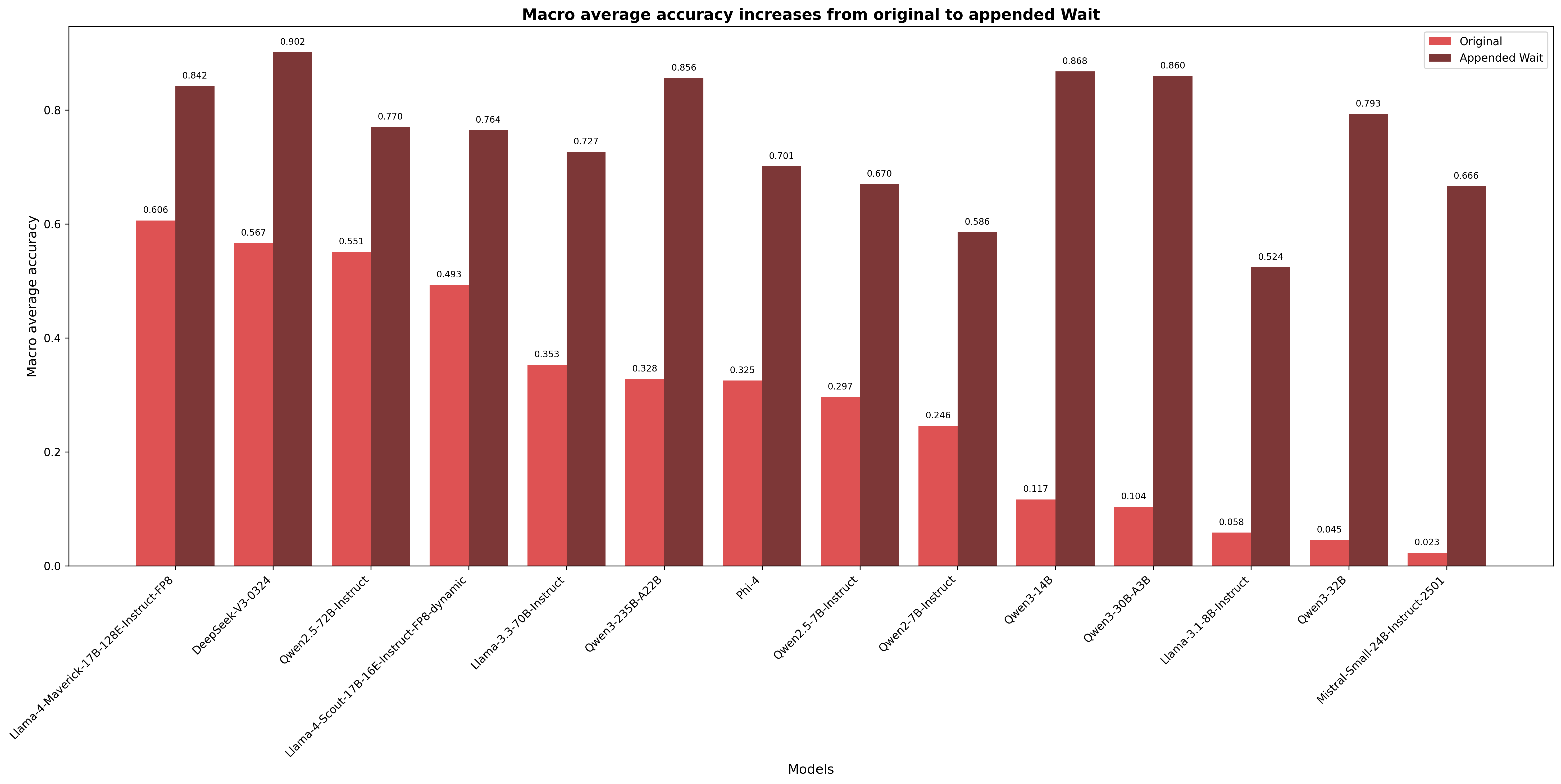}
    \caption{Macro average accuracy by non-reasoning model increases from original to appended ``\textit{Wait}"}
    \label{fig:error_injection_model_macro_averages_non_reasoning_no_wait_vs_wait}
\end{figure}

\begin{table}[ht]
    \centering
    % \footnotesize
    \begin{tabular}{llll}
        \toprule
Correction Markers & SCLI5 & GSM8K-SC & PRM800K-SC \\
\midrule
Internal Error (Baseline) & 0.499 (0\%) & 0.183 (0\%) & 0.200 (0\%) \\
External Error & 0.910 (+82.5\%) & 0.881 (+382.1\%) & 0.620 (+210.3\%) \\
``\textit{Wait}" & 0.957 (+91.9\%) & 0.796 (+335.1\%) & 0.504 (+152.0\%) \\
``\textit{But}" & 0.922 (+85.0\%) & 0.611 (+234.2\%) & 0.430 (+114.8\%) \\ 
``\textit{However}" & 0.897 (+79.8\%) & 0.602 (+229.0\%) & 0.438 (+119.3\%)  \\
        \bottomrule
    \end{tabular}
    \caption{Mean accuracy and relative change after appending various correction markers}
    \label{tab:variation_correction_markers}
\end{table}

\begin{figure}[ht]
    \centering
    \includegraphics[width=1.0\textwidth]{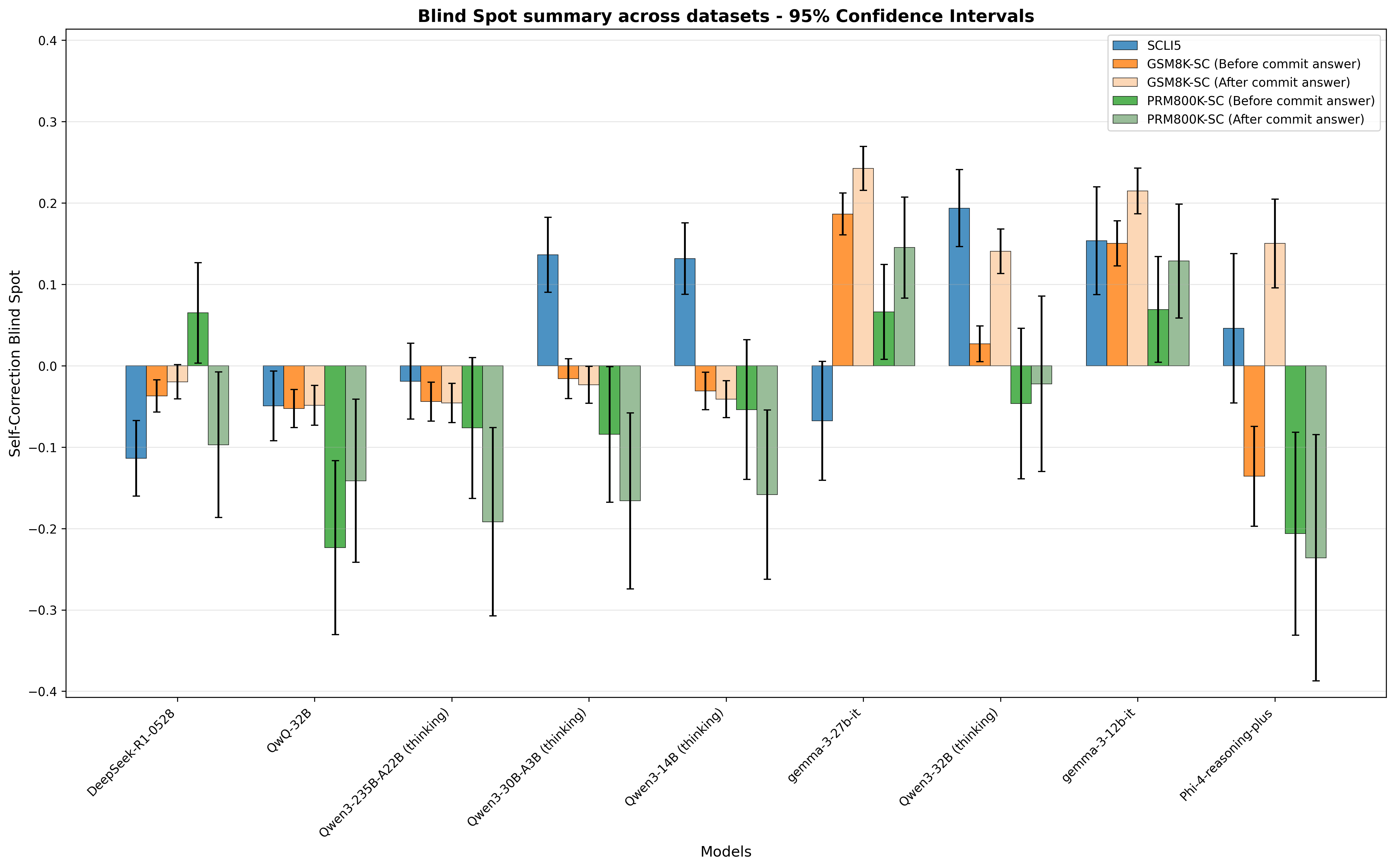}
    \caption{Self-Correction Blind Spot and 95\% confidence interval across reasoning models}
    \label{fig:blind_spot_summary_default_reasoning}
\end{figure}

\begin{table}[t]
    \centering
    \footnotesize
\begin{tabular}{lcccc}
\toprule
Condition & $\alpha=-1$ & $\alpha=-2$ & $\alpha=-3$ & $\alpha=-5$ \\
\midrule
\multicolumn{5}{l}{\textit{Llama-3.1-8B-Instruct / GSM8K-SC} \quad Internal ($\alpha{=}0$): 0.019, External: 0.652} \\
Real & 0.034 & 0.135 & 0.258 & 0.314 \\
Opposite & 0.006 & 0.000 & 0.000 & 0.001 \\
Random & 0.018 & 0.018 & 0.019 & 0.028 \\
$\Delta$(real$-$random) & \textbf{+0.016 $\pm$ 0.012} & \textbf{+0.118 $\pm$ 0.020} & \textbf{+0.239 $\pm$ 0.025} & \textbf{+0.286 $\pm$ 0.027} \\
\midrule
\multicolumn{5}{l}{\textit{Llama-3.1-8B-Instruct / SCLI5} \quad Internal ($\alpha{=}0$): 0.136, External: 0.797} \\
Real & 0.091 & 0.371 & 0.486 & 0.273 \\
Opposite & 0.038 & 0.000 & 0.000 & 0.000 \\
Random & 0.035 & 0.000 & 0.000 & 0.122 \\
$\Delta$(real$-$random) & \textbf{+0.056 $\pm$ 0.040} & \textbf{+0.371 $\pm$ 0.056} & \textbf{+0.486 $\pm$ 0.058} & \textbf{+0.150 $\pm$ 0.064} \\
\midrule
\multicolumn{5}{l}{\textit{Llama-3.1-8B-Instruct / PRM800K-SC} \quad Internal ($\alpha{=}0$): 0.020, External: 0.181} \\
Real & 0.018 & 0.062 & 0.134 & 0.123 \\
Opposite & 0.020 & 0.013 & 0.007 & 0.000 \\
Random & 0.020 & 0.020 & 0.022 & 0.018 \\
$\Delta$(real$-$random) & -0.002 $\pm$ 0.018 & \textbf{+0.042 $\pm$ 0.026} & \textbf{+0.112 $\pm$ 0.034} & \textbf{+0.105 $\pm$ 0.033} \\
\midrule
\multicolumn{5}{l}{\textit{Qwen2.5-7B-Instruct / GSM8K-SC} \quad Internal ($\alpha{=}0$): 0.190, External: 0.895} \\
Real & 0.256 & 0.272 & 0.314 & 0.608 \\
Opposite & 0.197 & 0.181 & 0.163 & 0.276 \\
Random & 0.226 & 0.216 & 0.191 & 0.192 \\
$\Delta$(real$-$random) & +0.030 $\pm$ 0.033 & \textbf{+0.056 $\pm$ 0.033} & \textbf{+0.123 $\pm$ 0.033} & \textbf{+0.416 $\pm$ 0.034} \\
\midrule
\multicolumn{5}{l}{\textit{Qwen2.5-7B-Instruct / SCLI5} \quad Internal ($\alpha{=}0$): 0.559, External: 0.682} \\
Real & 0.577 & 0.612 & 0.633 & 0.664 \\
Opposite & 0.538 & 0.608 & 0.643 & 0.724 \\
Random & 0.559 & 0.556 & 0.598 & 0.643 \\
$\Delta$(real$-$random) & +0.017 $\pm$ 0.081 & +0.056 $\pm$ 0.081 & +0.035 $\pm$ 0.080 & +0.021 $\pm$ 0.078 \\
\midrule
\multicolumn{5}{l}{\textit{Qwen2.5-7B-Instruct / PRM800K-SC} \quad Internal ($\alpha{=}0$): 0.141, External: 0.576} \\
Real & 0.156 & 0.179 & 0.205 & 0.266 \\
Opposite & 0.143 & 0.123 & 0.100 & 0.107 \\
Random & 0.147 & 0.163 & 0.145 & 0.143 \\
$\Delta$(real$-$random) & +0.009 $\pm$ 0.047 & +0.016 $\pm$ 0.049 & \textbf{+0.060 $\pm$ 0.050} & \textbf{+0.123 $\pm$ 0.052} \\
\bottomrule
\end{tabular}
    \caption{
\textbf{Activation steering along the conversational-role direction causally increases self-correction.} For each model and dataset, we report mean accuracy under steering at strength $\alpha$ along the direction (\emph{Real}), its negation (\emph{Opposite}), and a norm-matched random direction control (\emph{Random}). Direction computed on GSM8K-SC, cross-applied to SCLI5 and PRM800K-SC. $\Delta(\text{real} - \text{random})$ separates the conversational-role effect from generic perturbation: it is significant at $\alpha = -3$ for all model–dataset pairs except Qwen2.5-7B-Instruct / SCLI5, where the small baseline blind spot (18\%) leaves little headroom to steer. At $\alpha = -5$ the controls no longer sit at the internal baseline on some pairs (Section \ref{sec:mechanistic}). Bold indicates $\Delta$ significantly above zero (95\% CI excludes zero)}

    \label{tab:steering_statsig}
\end{table}

% \clearpage
% \input{appendix/maths_supplement}
\clearpage
\section{Sensitivity analysis}
\label{sensitivity_analysis}
\subsection{Result of different temperature}
Apart from using models' most confident prediction, we use temperature of 0.0 for 3 reasons:
\begin{itemize}
    \item More deterministic\footnote{Temperature of 0.0 will not generate fully deterministic results due to finite precision.} output eliminates sampling variance as a confounding factor.
    \item It enables standardized comparison across models with different temperature calibrations.
    \item \citet{renze-2024-effect} suggests different temperatures do not have a statistically significant impact on LLM performance in problem-solving tasks.
\end{itemize}    

\begin{table}[ht!]
\begin{threeparttable}
    \centering
    \footnotesize
    \begin{tabular}{llll}
        \toprule
Model & SCLI5 & GSM8K-SC & PRM800K-SC \\
\midrule
Llama-4-Maverick-17B-128E-Instruct-FP8 \citep{llama4} & 0.948 {\scriptsize ± 0.026} & 0.416 {\scriptsize ± 0.027} & 0.455 {\scriptsize ± 0.046} \\
DeepSeek-V3-0324 \citep{deepseekai2025deepseekv3technicalreport} & 0.825 {\scriptsize ± 0.044} & 0.399 {\scriptsize ± 0.026} & 0.475 {\scriptsize ± 0.046} \\
Qwen2.5-72B-Instruct \citep{qwen2025qwen25technicalreport} & 0.92 {\scriptsize ± 0.032} & 0.58 {\scriptsize ± 0.027} & 0.154 {\scriptsize ± 0.033} \\
Llama-4-Scout-17B-16E-Instruct \citep{llama4} & 0.976 {\scriptsize ± 0.018} & 0.24 {\scriptsize ± 0.023} & 0.263 {\scriptsize ± 0.041} \\
Llama-3.3-70B-Instruct \citep{llama3_3} & 0.538 {\scriptsize ± 0.058} & 0.275 {\scriptsize ± 0.024} & 0.246 {\scriptsize ± 0.04} \\
Qwen3-235B-A22B \footnotemark[1] \citep{yang2025qwen3technicalreport} & 0.563 {\scriptsize ± 0.058} & 0.073 {\scriptsize ± 0.014} & 0.348 {\scriptsize ± 0.044} \\
phi-4 \citep{abdin2024phi4technicalreport} & 0.808 {\scriptsize ± 0.046} & 0.076 {\scriptsize ± 0.014} & 0.092 {\scriptsize ± 0.027} \\
Qwen2.5-7B-Instruct  \citep{qwen2025qwen25technicalreport} & 0.559 {\scriptsize ± 0.058} & 0.19 {\scriptsize ± 0.021} & 0.141 {\scriptsize ± 0.032} \\
Qwen2-7B-Instruct \citep{yang2024qwen2technicalreport} & 0.601 {\scriptsize ± 0.057} & 0.078 {\scriptsize ± 0.014} & 0.058 {\scriptsize ± 0.022} \\
Qwen3-14B \footnotemark[1] \citep {yang2025qwen3technicalreport} & 0.004 {\scriptsize ± 0.007} & 0.092 {\scriptsize ± 0.016} & 0.254 {\scriptsize ± 0.04} \\
Qwen3-30B-A3B \footnotemark[1] \citep{yang2025qwen3technicalreport} & 0.056 {\scriptsize ± 0.027} & 0.061 {\scriptsize ± 0.013} & 0.194 {\scriptsize ± 0.037} \\
Llama-3.1-8B-Instruct \citep{grattafiori2024llama3herdmodels} & 0.136 {\scriptsize ± 0.04} & 0.019 {\scriptsize ± 0.007} & 0.02 {\scriptsize ± 0.013} \\
Qwen3-32B \footnotemark[1] \citep{yang2025qwen3technicalreport} & 0.004 {\scriptsize ± 0.007} & 0.05 {\scriptsize ± 0.012} & 0.083 {\scriptsize ± 0.026} \\
Mistral-Small-24B-Instruct-2501 \citep{mistral-small-3} & 0.042 {\scriptsize ± 0.023} & 0.011 {\scriptsize ± 0.006} & 0.016 {\scriptsize ± 0.012} \\
        \bottomrule
    \end{tabular}
    \begin{tablenotes}[flushleft]
    \small
    \item[1] Qwen3 series models use non-thinking mode.
    \end{tablenotes}
    \caption{Mean accuracy and 95\% confidence interval of models at temperature 0.0}
    \label{tab:mean-accuracy}
\end{threeparttable}
\end{table}

We also report results using a temperature of 0.6 below and the result does not change our conclusion.

\begin{table}[ht!]
    \centering
    \footnotesize
    \begin{tabular}{llll}
        \toprule
Model & SCLI5 & GSM8K-SC & PRM800K-SC \\
\midrule
Llama-4-Maverick-17B-128E-Instruct-FP8 & 0.954 {\scriptsize ± 0.024} & 0.424 {\scriptsize ± 0.027} & 0.469 {\scriptsize ± 0.046} \\
DeepSeek-V3-0324 & 0.874 {\scriptsize ± 0.039} & 0.42 {\scriptsize ± 0.027} & 0.504 {\scriptsize ± 0.046} \\
Qwen2.5-72B-Instruct & 0.902 {\scriptsize ± 0.035} & 0.574 {\scriptsize ± 0.027} & 0.165 {\scriptsize ± 0.034} \\
Llama-4-Scout-17B-16E-Instruct & 0.976 {\scriptsize ± 0.018} & 0.248 {\scriptsize ± 0.023} & 0.272 {\scriptsize ± 0.041} \\
Llama-3.3-70B-Instruct & 0.496 {\scriptsize ± 0.058} & 0.273 {\scriptsize ± 0.024} & 0.243 {\scriptsize ± 0.04} \\
Qwen3-235B-A22B & 0.57 {\scriptsize ± 0.057} & 0.091 {\scriptsize ± 0.016} & 0.4 {\scriptsize ± 0.045} \\
phi-4 & 0.794 {\scriptsize ± 0.047} & 0.093 {\scriptsize ± 0.016} & 0.116 {\scriptsize ± 0.03} \\
Qwen2.5-7B-Instruct & 0.563 {\scriptsize ± 0.058} & 0.183 {\scriptsize ± 0.021} & 0.127 {\scriptsize ± 0.031} \\
Qwen2-7B-Instruct & 0.601 {\scriptsize ± 0.057} & 0.071 {\scriptsize ± 0.014} & 0.065 {\scriptsize ± 0.023} \\
Qwen3-14B & 0.007 {\scriptsize ± 0.01} & 0.101 {\scriptsize ± 0.016} & 0.27 {\scriptsize ± 0.041} \\
Qwen3-30B-A3B & 0.108 {\scriptsize ± 0.036} & 0.07 {\scriptsize ± 0.014} & 0.232 {\scriptsize ± 0.039} \\
Qwen3-32B & 0.038 {\scriptsize ± 0.022} & 0.068 {\scriptsize ± 0.014} & 0.105 {\scriptsize ± 0.028} \\
Llama-3.1-8B-Instruct & 0.182 {\scriptsize ± 0.045} & 0.025 {\scriptsize ± 0.008} & 0.022 {\scriptsize ± 0.014} \\
Mistral-Small-24B-Instruct-2501 & 0.122 {\scriptsize ± 0.038} & 0.02 {\scriptsize ± 0.008} & 0.038 {\scriptsize ± 0.018} \\
        \bottomrule
    \end{tabular}
    \caption{Mean accuracy and 95\% confidence interval of models at temperature 0.6}
    \label{tab:mean-accuracy-06}
\end{table}

\newpage
\subsection{PRM800K-SC result in 4,096 token budget}
To ensure a fair comparison between internal and external error correction, and across models, we maintain a fixed token budget of 1,024 across all conditions. This design choice partly isolates self-correction capabilities from the effect of test time compute, providing a more rigorous test of the blind spot phenomenon. We also report our results of PRM800K-SC with a fixed token budget of 4,096 below, which does not change our conclusion. We do not report the result of SCLI5 and GSM8K-SC as the ratio of model responses exceeding 1,024 tokens is immaterial. 

\begin{table}[ht!]
    \centering
    \footnotesize
    \begin{tabular}{lllllll}
        \toprule
Model & \multicolumn{2}{c}{External Error} & \multicolumn{2}{c}{Internal Error} & \multicolumn{2}{c}{Appending ``\textit{Wait}"} \\
Compute budget & 1,024 & 4,096 & 1,024 & 4,096 & 1,024 & 4,096 \\
\midrule
Llama-4-Maverick-17B-128E-Instruct-FP8 & 0.71 & 0.721 & 0.455 & 0.458 & 0.67 & 0.676 \\
DeepSeek-V3-0324 & 0.775 & 0.938 & 0.475 & 0.509 & 0.772 & 0.821 \\
Qwen2.5-72B-Instruct & 0.612 & 0.614 & 0.154 & 0.161 & 0.438 & 0.449 \\
Llama-4-Scout-17B-16E-Instruct & 0.58 & 0.578 & 0.263 & 0.257 & 0.545 & 0.542 \\
Llama-3.3-70B-Instruct & 0.359 & 0.366 & 0.246 & 0.257 & 0.46 & 0.469 \\
Qwen3-235B-A22B & 0.786 & 0.806 & 0.348 & 0.368 & 0.705 & 0.732 \\
phi-4 & 0.714 & 0.719 & 0.092 & 0.092 & 0.328 & 0.337 \\
Qwen2.5-7B-Instruct & 0.576 & 0.569 & 0.141 & 0.141 & 0.442 & 0.444 \\
Qwen2-7B-Instruct & 0.658 & 0.65 & 0.058 & 0.058 & 0.324 & 0.333 \\
Qwen3-14B & 0.705 & 0.743 & 0.254 & 0.268 & 0.696 & 0.746 \\
Qwen3-30B-A3B & 0.779 & 0.817 & 0.194 & 0.19 & 0.683 & 0.712 \\
Qwen3-32B & 0.754 & 0.781 & 0.083 & 0.085 & 0.527 & 0.522 \\
Llama-3.1-8B-Instruct & 0.181 & 0.183 & 0.02 & 0.02 & 0.194 & 0.203 \\
Mistral-Small-24B-Instruct-2501 & 0.496 & 0.498 & 0.016 & 0.016 & 0.27 & 0.277 \\
        \bottomrule
    \end{tabular}
    \caption{Mean accuracy of models in PRM800K-SC at different compute budget}
    \label{tab:diff_budget_prm800k_sc}
\end{table}

\subsection{Choice of LLM judge}
\label{sensitivity_analysis_llm_judge}
We compare Gemini 2.5 Flash with Claude Sonnet 4.6 and GPT-5.4 on 238 samples. Three-way agreement is 95.0\%, demonstrating robustness of LLM judge.
\begin{table}[ht]
    \centering
    \begin{tabular}{lll}
        \toprule
Model Comparison & Agreement & Cohen's kappa \\
\midrule
Gemini 2.5 Flash vs Claude Sonnet 4.6 & 0.979 & 0.953 \\ 
Gemini 2.5 Flash vs GPT-5.4 & 0.954 & 0.899 \\
Claude Sonnet 4.6 vs GPT-5.4 & 0.967 & 0.928 \\
        \bottomrule
    \end{tabular}
    \caption{Inter-judge agreement}
    \label{tab:inter_judge_argeement}
\end{table}

\clearpage

\clearpage
\section{Prompt}
\subsection{Generating GSM8K-SC}
\label{prompt_for_gsm8k_sc}
\begin{figure}[ht!]
\centering
\begin{minipage}[t]{1.0\textwidth}
\begin{tcolorbox}[colback=gray!10, colframe=gray!50, rounded corners, arc=15pt]
\begin{lstlisting}[breaklines=true]
from pydantic import BaseModel


class ReasoningWithMistake(BaseModel):
    reasoning_steps_with_one_mistake: List[str]
    mistake_step: int
    type_of_mistake: str
    description_of_mistake: str
    incorrect_answer: str
\end{lstlisting}
\end{tcolorbox}
\end{minipage}
\begin{minipage}[t]{1.0\textwidth}
\begin{tcolorbox}[colback=gray!10, colframe=gray!50, rounded corners, arc=15pt]
\begin{lstlisting}[breaklines=true]
You are a helpful assistant that follow instructions. Output in JSON format.
\end{lstlisting}
\end{tcolorbox}
\end{minipage}

\begin{minipage}[t]{1.0\textwidth}
\begin{tcolorbox}[colback=gray!10, colframe=gray!50, rounded corners, arc=15pt]
\begin{lstlisting}[breaklines=true]
<question> 
{question} 
</question>

<reasoning_steps> 
{reasoning_steps}
</reasoning_steps>

<answer> 
{answer}
</answer>

<type_of_mistake>
{error_type}: {error_description}
</type_of_mistake>

You task is to introduce one mistake in step {mistake_step} in <reasoning_steps> and arrive at an answer different from <answer>. 
You will output: 
- <reasoning_steps> with mistake
- the step that contains the mistake
- type of the mistake
- description of the mistake
- incorrect answer
\end{lstlisting}
\end{tcolorbox}
\end{minipage}
\hfill

\caption{Output schema, system prompt and prompt for generating GSM8K-SC dataset}
\label{fig:generate-gsm8k-sc-prompt}
\end{figure}

\begin{figure}[ht!]
\centering
\begin{minipage}[t]{1.0\textwidth}
\begin{tcolorbox}[colback=gray!10, colframe=gray!50, rounded corners, arc=15pt]
\begin{lstlisting}[breaklines=true]
from pydantic import BaseModel


class Calculation(BaseModel):
    incorrect_answer: float
\end{lstlisting}
\end{tcolorbox}
\end{minipage}
\begin{minipage}[t]{1.0\textwidth}
\begin{tcolorbox}[colback=gray!10, colframe=gray!50, rounded corners, arc=15pt]
\begin{lstlisting}[breaklines=true]
You are a helpful assistant that follow instructions. Output in JSON format.
\end{lstlisting}
\end{tcolorbox}
\end{minipage}

\begin{minipage}[t]{1.0\textwidth}
\begin{tcolorbox}[colback=gray!10, colframe=gray!50, rounded corners, arc=15pt]
\begin{lstlisting}[breaklines=true]
<question>
{question}
</question>

<description_of_mistake>
{description_of_mistake}
</description_of_mistake>

<incorrect_reasoning>
{incorrect_reasoning}
</incorrect_reasoning>

You are given a question, a description of the mistake, and the resulting incorrect reasoning.
Your task is to follow the incorrect reasoning to arrive at the incorrect answer.
\end{lstlisting}
\end{tcolorbox}
\end{minipage}
\hfill

\caption{Output schema, system prompt and prompt for validating GSM8K-SC dataset}
\label{fig:validate-gsm8k-sc-prompt}
\end{figure}

\clearpage
\subsection{Automatic Evaluation}
\label{prompt_automatic_evaluation}

\begin{figure}[ht!]
\centering
\begin{minipage}[t]{1.0\textwidth}
\begin{tcolorbox}[colback=gray!10, colframe=gray!50, rounded corners, arc=15pt]
\begin{lstlisting}[breaklines=true]
from pydantic import BaseModel


class Evaluation(BaseModel):
    is_correct_answer: bool
    is_correct_reasoning: bool
    is_backtracking: bool
    is_admitting_mistake: bool
\end{lstlisting}
\end{tcolorbox}
\end{minipage}
\begin{minipage}[t]{1.0\textwidth}
\begin{tcolorbox}[colback=gray!10, colframe=gray!50, rounded corners, arc=15pt]
\begin{lstlisting}[breaklines=true]
You are a helpful assistant that follow instructions. Output in JSON format.
\end{lstlisting}
\end{tcolorbox}
\end{minipage}
\hfill
\caption{Output schema and system prompt for automatic evaluation of error in model}
\label{fig:auto-evaluation-model1}
\end{figure}

\begin{figure}[ht!]
\centering
\begin{minipage}[t]{1.0\textwidth}
\begin{tcolorbox}[colback=gray!10, colframe=gray!50, rounded corners, arc=15pt]
\begin{lstlisting}[breaklines=true]
# FOR SCLI5
<question>
{question}
</question>

<golden_answer>
{golden_answer}
</golden_answer>

<given_wrong_answer>
{given_wrong_answer}
</given_wrong_answer>

<completion_from_model>
{completion_from_model}
</completion_from_model>

The model was provided a wrong answer <given_wrong_answer>, and is required to self correct and arrive at the golden answer.
You have to assess if <completion_from_model> :
- has backtracked
- has finally provided the answer that matches the <golden_answer>
- has provided correct reasoning
- has admitted the mistake.
\end{lstlisting}
\end{tcolorbox}
\end{minipage}
\begin{minipage}[t]{1.0\textwidth}
\begin{tcolorbox}[colback=gray!10, colframe=gray!50, rounded corners, arc=15pt]
\begin{lstlisting}[breaklines=true]
# FOR GSM8K-SC AND PRM800K-SC
<question>
{question}
</question>

<golden_answer>
{golden_answer}
</golden_answer>

<given_wrong_reasoning>
{given_wrong_reasoning}
</given_wrong_reasoning>

<completion_from_model>
{completion_from_model}
</completion_from_model>

The model was provided a wrong step in the reasoning process <given_wrong_reasoning>, and is required to self correct and arrive at the golden answer.
You have to assess if <completion_from_model> :
- has backtracked
- has finally provided the answer that matches the <golden_answer>
- has provided correct reasoning
- has admitted the mistake.
\end{lstlisting}
\end{tcolorbox}
\end{minipage}
\hfill
\caption{Prompt for automatic evaluation of error in model}
\label{fig:auto-evaluation-model2}
\end{figure}

\begin{figure}[ht!]
\centering
\begin{minipage}[t]{1.0\textwidth}
\begin{tcolorbox}[colback=gray!10, colframe=gray!50, rounded corners, arc=15pt]
\begin{lstlisting}[breaklines=true]
from pydantic import BaseModel


class EvaluationErrorInUser(BaseModel):
    is_correct_answer: bool
    is_correct_critique: bool
\end{lstlisting}
\end{tcolorbox}
\end{minipage}
\begin{minipage}[t]{1.0\textwidth}
\begin{tcolorbox}[colback=gray!10, colframe=gray!50, rounded corners, arc=15pt]
\begin{lstlisting}[breaklines=true]
You are a helpful assistant that follow instructions. Output in JSON format.
\end{lstlisting}
\end{tcolorbox}
\end{minipage}
\hfill
\caption{Output schema and system prompt for automatic evaluation of error in user}
\label{fig:auto-evaluation-user1}
\end{figure}

\begin{figure}[ht!]
\centering
\begin{minipage}[t]{1.0\textwidth}
\begin{tcolorbox}[colback=gray!10, colframe=gray!50, rounded corners, arc=15pt]
\begin{lstlisting}[breaklines=true]
# FOR SCLI5
<question_and_user_answer>
{question_and_user_answer}
</question_and_user_answer>

<golden_answer>
{golden_answer}
</golden_answer>

<response_from_model>
{response_from_model}
</response_from_model>

The model was provided with <question_and_user_answer> from user.
You have to assess if <response_from_model> :
- contains correct answer that matches the <golden_answer>
- contains correct critique of <question_and_user_answer>
\end{lstlisting}
\end{tcolorbox}
\end{minipage}
\begin{minipage}[t]{1.0\textwidth}
\begin{tcolorbox}[colback=gray!10, colframe=gray!50, rounded corners, arc=15pt]
\begin{lstlisting}[breaklines=true]
# FOR GSM8K-SC AND PRM800K-SC
<question_and_user_reasoning>
{question_and_user_reasoning}
</question_and_user_reasoning>

<golden_answer>
{golden_answer}
</golden_answer>

<response_from_model>
{response_from_model}
</response_from_model>

The model was provided with <question_and_user_reasoning> from user.
You have to assess if <response_from_model> :
- contains correct answer that matches the <golden_answer>
- contains correct critique of <question_and_user_reasoning>
\end{lstlisting}
\end{tcolorbox}
\end{minipage}
\hfill
\caption{Prompt for automatic evaluation of error in user}
\label{fig:auto-evaluation-user2}
\end{figure}

\clearpage
\section{Example of Self-Correction Bench}
\subsection{SCLI5}
\label{example_scli5}
\begin{figure}[ht!]
\centering
\begin{minipage}[t]{1.0\textwidth}
\begin{tcolorbox}[colback=gray!10, colframe=gray!50, rounded corners, arc=15pt]
\begin{lstlisting}[breaklines=true,basicstyle=\small]
# error in user prompt
[
    {
        "content": "What is the answer of 20 + 1? The answer is 22.",
        "role": "user"
    }
]
# error in model response
[
    {
        "content": "What is the answer of 20 + 1?",
        "role": "user"
    },
    {
        "content": "The answer is 22.",
        "role": "assistant"
    }
]
\end{lstlisting}
\end{tcolorbox}
\end{minipage}
\end{figure}

\subsection{GSM8K-SC}
\label{example_gsm8k_sc}
\begin{figure}[ht!]
\centering
\begin{minipage}[t]{1.0\textwidth}
\begin{tcolorbox}[colback=gray!10, colframe=gray!50, rounded corners, arc=15pt]
\begin{lstlisting}[breaklines=true,basicstyle=\small]
# error in user prompt
[
    {
        "content": "Harry slept 9 hours last night. His friend James slept only 2/3 of what Harry slept. How many more hours did Harry sleep than James? James slept 9 hours / 2/3 = 13.5 hours. So, Harry slept 9 - 13.5 = -4.5 hours more than James.",
        "role": "user"
    }
]
# error in model response
[
    {
        "content": "Harry slept 9 hours last night. His friend James slept only 2/3 of what Harry slept. How many more hours did Harry sleep than James?",
        "role": "user"
    },
    {
        "content": "James slept 9 hours / 2/3 = 13.5 hours. So, Harry slept 9 - 13.5 = -4.5 hours more than James.",
        "role": "assistant"
    }
]
\end{lstlisting}
\end{tcolorbox}
\end{minipage}
\end{figure}

\clearpage
\subsection{PRM800K-SC}
\label{example_prm800k_sc}
\begin{figure}[ht!]
\centering
\begin{minipage}[t]{1.0\textwidth}
\begin{tcolorbox}[colback=gray!10, colframe=gray!50, rounded corners, arc=15pt]
\begin{lstlisting}[breaklines=true,basicstyle=\small]
# error in user prompt
[
    {
        "content": "A Senate committee has 5 Democrats, 5 Republicans, and 1 Independent. In how many ways can they sit around a circular table if all the members of each party all sit next to each other? (Two seatings are considered equivalent if one is a rotation of the other.) I notice that there are three groups of people: Democrats, Republicans, and Independent. I wonder how many ways I can arrange these groups around the circle, ignoring the order within each group for now. I recall that the number of ways to arrange n distinct objects around a circle is (n-1)!, since we can fix one object and then permute the rest. (Truncated due to size.) So, there are 5!/2 = 60 ways to order the Democrats around a circle. Similarly, there are 60 ways to order the Republicans around a circle. For the Independent, there is only one way to order him or her around a circle. So, to get the total number of ways to seat the committee, I need to multiply the number of ways to arrange the groups by the number of ways to order each group. That gives me 2 x 60 x 60 x 1 = 7200 ways.",
        "role": "user"
    }
]
# error in model response
[
    {
        "content": "A Senate committee has 5 Democrats, 5 Republicans, and 1 Independent. In how many ways can they sit around a circular table if all the members of each party all sit next to each other? (Two seatings are considered equivalent if one is a rotation of the other.)",
        "role": "user"
    },
    {
        "content": "I notice that there are three groups of people: Democrats, Republicans, and Independent. I wonder how many ways I can arrange these groups around the circle, ignoring the order within each group for now. I recall that the number of ways to arrange n distinct objects around a circle is (n-1)!, since we can fix one object and then permute the rest. (Truncated due to size.) So, there are 5!/2 = 60 ways to order the Democrats around a circle. Similarly, there are 60 ways to order the Republicans around a circle. For the Independent, there is only one way to order him or her around a circle. So, to get the total number of ways to seat the committee, I need to multiply the number of ways to arrange the groups by the number of ways to order each group. That gives me 2 x 60 x 60 x 1 = 7200 ways.",
        "role": "assistant"
    }
]
\end{lstlisting}
\end{tcolorbox}
\end{minipage}
\end{figure}
\flushbottom

\end{document}